\documentclass[letterpaper, 10 pt, conference]{ieeeconf}  

\IEEEoverridecommandlockouts                              

\usepackage{iftex}
\ifLuaTeX
  \usepackage{fontspec}
\else
  \usepackage{times}
\fi
\usepackage{siunitx}
\usepackage{cite} 

\usepackage{graphicx} 
\usepackage{amsmath} 
\usepackage{amssymb}  
\usepackage{balance}
\usepackage{caption}
\usepackage{subfigure} 
\usepackage{tikz} 
\usepackage{pgfplots} 
\usepackage{float} 
\usepackage{pifont}
\pgfplotsset{
  compat=1.18,
  every axis/.append style={
    /tikz/.append style={font=\rmfamily},
  },
}
\usetikzlibrary{shapes.geometric, positioning, arrows.meta, patterns}
\usepgfplotslibrary{statistics}
\usepackage{hyperref}
\usepackage{multirow} 
\usepackage{booktabs} 
\usepackage[colorinlistoftodos]{todonotes}

\newcommand{\cmark}{\ding{51}}%
\newcommand{\xmark}{\ding{55}}%

\newcommand{\Tabref}[1]{Table~\ref{#1}}

\title{\LARGE \bf
TriSAR-3D and PaintFusion: \\ A Novel Trimodal LiDAR–RGB–Thermal Dataset and Method for \\ 3D Person Detection in Search and Rescue}

 \author{
 Markus Kramer$^{1,\ast}$,
 Jan Wachenfeld$^{1,\ast}$,
 and Oskar von Stryk$^{1}$
   \thanks{\raggedright $^{1}$The authors are with the SIM Group, Technical University of Darmstadt, Germany. Email: \texttt{kramer@sim.tu-darmstadt.de}, \texttt{jan.wachenfeld@stud.tu-darmstadt.de}, \texttt{stryk@sim.tu-darmstadt.de}.}%
   \thanks{\raggedright $^{\ast}$Markus Kramer and Jan Wachenfeld contributed equally to this work.}%
   \thanks{\raggedright Research presented in this paper has been supported in parts by the LOEWE initiative (Hesse, Germany) within the emergenCITY center [LOEWE/1/12/519/03/05.001(0016)/72] and by the Federal Ministry of Research, Technology and Space (BMFTR) within the projects DRZ (grant no. 13N16475) and KIARA PRO (grant no. 13N17765), and within the Hessian Drone Competence Centre Kassel Airport (HDKZ), funded by the European Regional Development Fund (ERDF), the Hessian Ministry for Digitalization and Innovation (Distr@l), and the Technical University of Darmstadt.}%
}

\begin{document}
\maketitle

\thispagestyle{empty}
\pagestyle{empty}

\begin{abstract}
    Reliable 3D person detection is important for rescue robotics to give first responders situational awareness. 
     Yet most multimodal 3D detection research targets automotive scenarios in which pedestrians mostly appear in a narrow range of upright, near-road poses. It is therefore unclear how much of the reported performance transfers to rescue settings, as a person may be lying, partially occluded, or in unstructured indoor and outdoor environments. To investigate this gap, we introduce TriSAR-3D, a multimodal person detection dataset recorded with a high-resolution 128-channel LiDAR, an RGB camera, and a thermal camera. It captures varied poses in indoor as well as outdoor locations at different times of day to study their influence on 3D object detector performance. Furthermore, we propose a novel trimodal 3D detection pipeline, PaintFusion, that fuses PointPainting-based evidence using a simple MLP. We show that the derived method reaches 65.4 AP@50 and thereby outperforms strong LiDAR-only (CenterPoint, 53.4), RGB+LiDAR (BEVFusion, 60.5), and adapted trimodal (Tri-BEVFusion, 60.7) baselines on the dataset. At the same time, it is substantially faster than other fusion-based methods such as BEVFusion, with a per-frame latency of \SI{51.6}{ms} compared to \SI{101}{ms}.
    Additionally, we test how the performance of different 3D object detectors generalizes to unseen pose categories. Removing lying instances from training data collapses the lying-pose AP@50 to near zero across all evaluated detectors. The results indicate that dataset diversity exerts a large effect on 3D object detection performance. The dataset and code are available at \url{https://github.com/MarkusKramer1/3D-Trimodal-People-Detection}.
\end{abstract}

\section{INTRODUCTION}

In search-and-rescue robotics, the objective of a mission is often to locate people in the scene, which may be indoors or outdoors, under lighting conditions that range from daylight to complete darkness. 3D detection models can contribute to such missions. While a 2D detection reveals only that a person is present, a 3D detection localizes that person precisely in the scene.
\begin{figure}[t!]
  \centering
  \includegraphics[width=\linewidth]{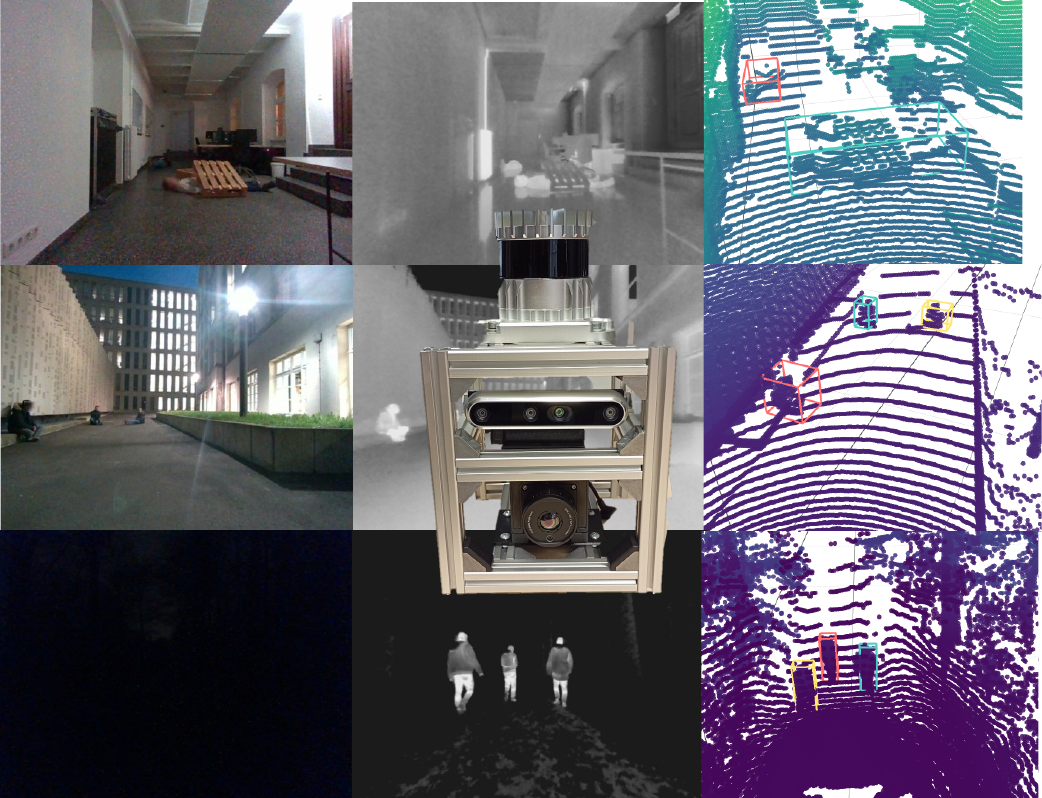}
  \caption{TriSAR-3D: synchronized RGB, thermal, and LiDAR recordings of people in varied poses, indoors and outdoors, by day and at night.}
  \label{fig:teaser}
  \vspace{-10pt}
\end{figure}

Most progress in 3D object detection is made on autonomous-driving datasets like KITTI~\cite{geigerVisionMeetsRobotics2013} and nuScenes~\cite{caesarNuScenesMultimodalDataset2020a}, which pair a LiDAR with RGB cameras and include pedestrians among many other categories. In these datasets, people appear almost always upright and near the road. Rescue robotics requires a model that works outside of this regime, especially for unusual poses and in the dark. 

A LiDAR provides accurate geometric measurements, but its point density decreases with range. Thus distant people return too few points to distinguish them from other structures. An RGB camera provides semantic cues for these cases, but depends on ambient light and performs poor in the dark. A thermal camera lacks this shortcoming: a body emits heat and thus appears as a warm region, unimpeded by the absence of illumination. Fusion of a LiDAR with an RGB and a thermal camera thus covers a wider range of conditions.

\begin{figure*}[!t]
    \centering
    \begin{minipage}[t]{0.62\textwidth}
        \centering
        \footnotesize
        \setlength{\tabcolsep}{2.5pt}
        \renewcommand{\arraystretch}{1.05}
        \begin{tabular}{@{}l|ccccc@{}}
                & STF~\cite{bijelicSeeingFogSeeing2020} & TriStar~\cite{stippelTrimodalDatasetRGB2024a} & Mid-3K~\cite{motaMultimodalHumanDetection2024a} & R-LiViT~\cite{mirlachRLiViTLiDARVisualThermalDataset2025} & \textbf{Ours} \\ 
            \hline
            Task & Auto.\ driving & Human seg. & Human det. & VRU det. & Human det. \\
            \hline
            \multicolumn{6}{@{}c@{}}{\textbf{Sensor setup}} \\
            \hline
            RGB cameras & \textbf{2} & 1 & 1 & 1 & 1 \\ 
            RGB resolution & \textbf{1920$\times$1024} & 640$\times$480 & 1280$\times$1024 & 1280$\times$720 & 640$\times$480 \\
            Thermal cameras & 1 & 1 & 1 & 1 & 1 \\
            Thermal resolution & 640$\times$480 & 640$\times$480 & \textbf{640$\times$512} & \textbf{640$\times$512} & 640$\times$480 \\
            Depth sensors & \textbf{2 LiDAR} & Depth camera & 1 LiDAR & \textbf{2 LiDAR} & 1 LiDAR\\
            Depth resolution & 64 channels & 640$\times$480 & 64 channels & 64/32 ch. & \textbf{128 channels} \\
            Frame rate [Hz] & \textbf{10} & 8.7 & - & 1.25/5 & \textbf{10} \\ 
            \hline
            \multicolumn{6}{@{}c@{}}{\textbf{Dataset statistics}} \\
            \hline
            Annotated frames & 13.5K & 15.6k & 3.1k & 10k & \textbf{19.7k} \\
            3D bboxes & 36.4k & - & - & \textbf{96k} & 52k\\
            2D RGB bboxes & 36.4k & - & 10.1k & 21k & \textbf{36.6k} \\
            2D thermal bboxes & - & - & 10.1k & 21k & \textbf{36.6k} \\
            2D RGB masks & - & 14k & - & - & \textbf{36.6k} \\
            2D thermal masks & - & 14k & - & - & \textbf{36.6k} \\
            \hline
            \multicolumn{6}{@{}c@{}}{\textbf{Scene composition}} \\
            \hline
            \multicolumn{6}{@{}l@{}}{\textit{Environment}} \\
            Day-time & \cmark & \cmark & \cmark & \cmark & \cmark \\
            Night-time & \cmark & \xmark & \xmark & \cmark & \cmark \\
            Indoor & \xmark & \cmark & \cmark & \xmark & \cmark \\
            Outdoor & \cmark & \xmark & \xmark & \cmark & \cmark \\
            \multicolumn{6}{@{}l@{}}{\textit{Poses}} \\
            Sitting & - & \cmark & - & - & \cmark \\
            Walking & - & \cmark & - & - & \cmark \\
            Lying & - & \cmark & - & - & \cmark \\
            Standing & - & \cmark & - & - & \cmark \\
        \end{tabular}
        \captionof{table}{Comparison of trimodal datasets with human annotations. TriSAR-3D is currently the only dataset that includes 3D bounding boxes and pose annotations in both indoor and outdoor scenarios during daytime and nighttime conditions. }
        \label{tab:dataset_compare}
    \end{minipage}%
    \hfill
    \begin{minipage}[t]{0.35\textwidth}
        \centering
        \definecolor{plotours}{HTML}{4E79A7}
        \definecolor{plottristar}{HTML}{F28E2B}
        \definecolor{plotrvlit}{HTML}{59A14F}
        \definecolor{plotstf}{HTML}{E15759}
        \definecolor{plotmid}{HTML}{CC00CC}
        \begin{tikzpicture}
            \begin{axis}[
                name=distchart,
                width=\linewidth,
                height=0.72\linewidth,
                ybar,
                x=1cm,
                bar width=0.08cm,
                enlarge x limits=0.15,
                symbolic x coords={0--10, 10--20, 20--30, 30--40, {40+}},
                xtick=data,
                xlabel={Distance [\si{m}]},
                ylabel={\# frames},
                ymin=0,
                ymax=32000,
                ytick={0,10000,20000,30000},
                yticklabel style={/pgf/number format/1000 sep={\,}, font=\scriptsize},
                tick label style={font=\scriptsize},
                label style={font=\footnotesize},
                legend style={
                    at={(0.5,1.03)},
                    anchor=south,
                    legend columns=3,
                    draw=none,
                    fill=none,
                    font=\scriptsize,
                    /tikz/every even column/.append style={column sep=0.3em},
                },
                single ybar legend,
            ]
                \addplot[fill=plotours, draw=plotours!70!black, line width=0.4pt] coordinates {
                    (0--10, 18544)
                    (10--20, 12416)
                    (20--30, 9181)
                    (30--40, 5493)
                    (40+, 6334)
                };
                \addplot[fill=plottristar, draw=plottristar!70!black, line width=0.4pt] coordinates {
                    (0--10, 14284)
                    (10--20, 0)
                    (20--30, 0)
                    (30--40, 0)
                    (40+, 0)
                };
                \addplot[fill=plotrvlit, draw=plotrvlit!70!black, line width=0.4pt] coordinates {
                    (0--10, 26779)
                    (10--20, 9770)
                    (20--30, 17019)
                    (30--40, 23404)
                    (40+, 19167)
                };
                \addplot[fill=plotstf, draw=plotstf!70!black, line width=0.4pt] coordinates {
                    (0--10, 847)
                    (10--20, 3553)
                    (20--30, 5033)
                    (30--40, 5566)
                    (40+, 25068)
                };
                \addplot[fill=plotmid, draw=plotmid!70!black, line width=0.4pt] coordinates {
                    (0--10, 3270)
                    (10--20, 3706)
                    (20--30, 1908)
                    (30--40, 1098)
                    (40+, 842)
                };
                \legend{Ours, TriStar$^{1}$, R-LiViT, STF, MID-3K$^{1}$}
            \end{axis}
        \end{tikzpicture}
        \vspace{-0.5em}
        \captionof{figure}{Distance distribution of datasets.\\[0.25ex]
        {\footnotesize $^{1}$\,Distance calculated from depth image.}}
        \label{fig:dataset_distance}

        \vspace{0.15em}
        \begin{tikzpicture}
            \begin{axis}[
                name=posechart,
                width=\linewidth,
                height=0.72\linewidth,
                ybar,
                bar width=0.22cm,
                enlarge x limits=0.16,
                symbolic x coords={Standing, Walking, Sitting, Lying},
                xtick=data,
                xticklabel style={font=\scriptsize},
                ylabel={\# frames},
                ymin=0,
                ymax=23000,
                ytick={0,5000,10000,15000,20000},
                yticklabel style={/pgf/number format/1000 sep={\,}, font=\scriptsize},
                tick label style={font=\scriptsize},
                label style={font=\footnotesize},
                legend style={
                    at={(0.5,1.02)},
                    anchor=south,
                    legend columns=2,
                    draw=none,
                    fill=none,
                    font=\scriptsize,
                },
                single ybar legend,
            ]
                \addplot[fill=plotours, draw=plotours!70!black, line width=0.4pt] coordinates {
                    (Standing, 2154)
                    (Walking, 21239)
                    (Sitting, 6713)
                    (Lying, 3047)
                };
                \addplot[fill=plottristar, draw=plottristar!70!black, line width=0.4pt] coordinates {
                    (Standing, 3901)
                    (Walking, 4699)
                    (Sitting, 3788)
                    (Lying, 404)
                };
                \legend{Ours, TriStar}
            \end{axis}
        \end{tikzpicture}
        \vspace{-0.5em}
        \captionof{figure}{Pose distribution of datasets which include pose annotations }
        \label{fig:dataset_pose}
    \end{minipage}
\end{figure*}

Two questions arise. How can a thermal camera be added into LiDAR--RGB fusion, and to what extent does it improve detection performance? And how do detectors fare on poses that are rare or absent in the training data? We study both with a new trimodal dataset and a controlled comparison.

In this work, we make three contributions:
\begin{itemize} 
  \item We introduce \emph{TriSAR-3D}, a trimodal person detection dataset recorded with a 128-channel LiDAR, an RGB camera, and a thermal camera. 
  \item We propose PaintFusion, a trimodal 3D detector that is the most accurate method we evaluate and achieves about half the latency of other fusion baselines.
  \item We study the effects of modality  choice and training set diversity on 3D detection performance.
\end{itemize}

\section{RELATED WORK}

We review existing multimodal datasets with 2D/3D person annotations and multimodal 3D detection methods.

\subsection{Datasets}
Public datasets that combine LiDAR, RGB, and thermal sensing with 2D and 3D annotations of people remain rare and differ strongly in task focus, sensor setup, and scene diversity.
Table~\ref{tab:dataset_compare} and Figs.~\ref{fig:dataset_distance} and \ref{fig:dataset_pose} summarize the publicly available trimodal datasets.
None of them targets search-and-rescue robotics directly, but they differ in how closely they match our setting.
Mid-3K~\cite{motaMultimodalHumanDetection2024a} is the closest in task focus, as it supports multimodal human detection with LiDAR, RGB, and thermal sensors mounted on a Clearpath Jackal robot.
TriStar~\cite{stippelTrimodalDatasetRGB2024a} is an indoor only RGB-Depth-Thermal dataset which provides dense 2D segmentation masks and explicit pose tags. However, it relies on a depth camera rather than a LiDAR and covers only indoor scenes.
R-LiViT~\cite{mirlachRLiViTLiDARVisualThermalDataset2025} offers a large set of 3D bounding boxes from a trimodal roadside setup focused on vulnerable road users (VRU) in traffic.
SeeingThroughFog~\cite{bijelicSeeingFogSeeing2020} (STF) is an automotive dataset with dual 64-channel LiDAR sensors and 3D labels. It has been recorded primarily for adverse-weather automated driving.

Across these datasets, annotation type and sensor resolution vary substantially.
To the best of our knowledge, only our dataset provides 3D bounding boxes, synchronized RGB and thermal instance masks, and pose tags for each person.
Its 128-channel LiDAR exceeds the LiDAR resolution of the other datasets and is better suited for detecting humans at longer range.
R-LiViT contains more 3D boxes overall, but without pose labels or indoor coverage.
TriStar and Mid-3K include only 2D bounding boxes or segmentation masks and lack 3D annotations.
In terms of scene composition, prior datasets typically cover either indoor or outdoor environments, whereas ours combines both.

The distribution plots in Figs.~\ref{fig:dataset_distance} and \ref{fig:dataset_pose} highlight two further gaps.
The distance histogram shows that TriStar annotations are concentrated within \SIrange{0}{10}{m}, consistent with the limited range of its depth sensor. Our dataset, in contrast, spans substantially farther distances. The STF dataset, on the other hand, mainly focuses on large-distance samples. 
Although TriStar includes all poses, the lying pose has only a few samples, making proper evaluation more difficult.

\begin{figure*}[!t]
    \centering
    \begin{minipage}[t]{0.49\textwidth}
        \centering
        \includegraphics[width=0.72\linewidth]{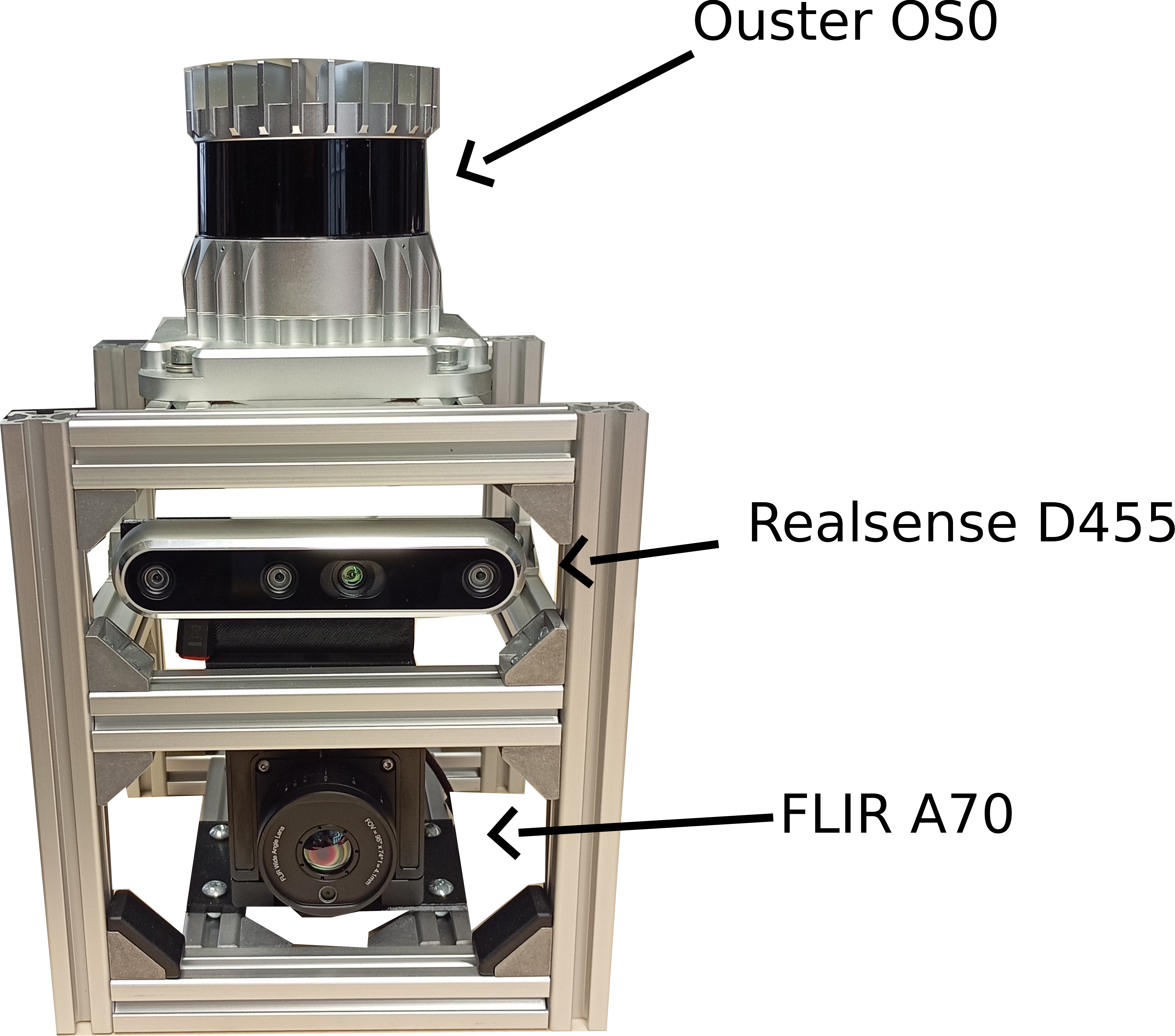}
        \captionof{figure}{Sensor-box with the LiDAR on top, the RGBD camera in the middle, and the thermal camera at the bottom.}
        \label{fig:sensorbox}
    \end{minipage}%
    \hfill
    \begin{minipage}[t]{0.49\textwidth}
        \centering
        \includegraphics[width=0.48\linewidth]{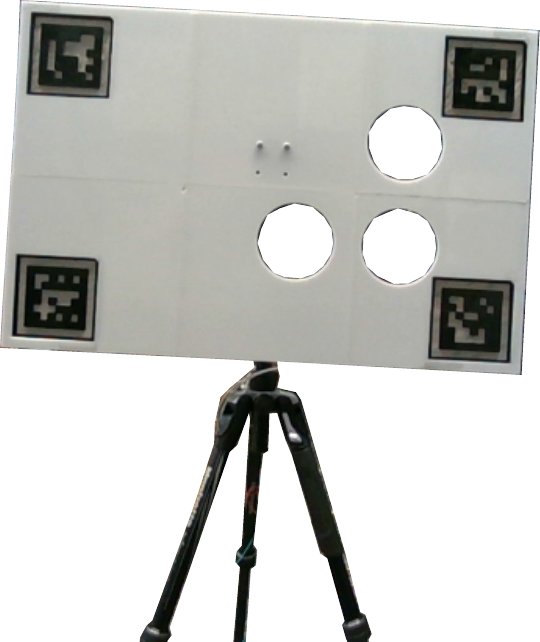}%
        \hfill
        \includegraphics[width=0.45\linewidth]{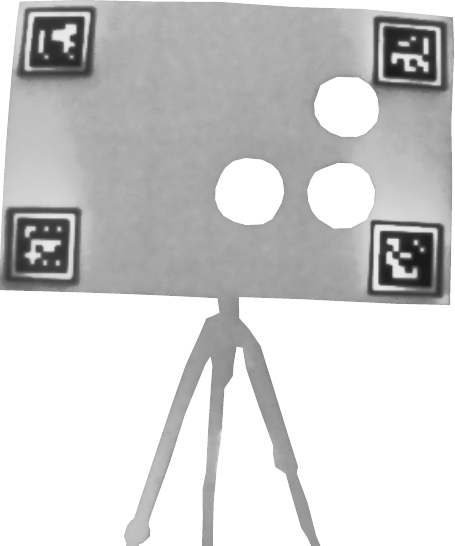}
        \captionof{figure}{Extrinsic calibration target in RGB image (left) and thermal image (right).}
        \label{fig:calib_target}
    \end{minipage}
\end{figure*}

\subsection{Multimodal Detection}
3D detection for mobile robots builds primarily on LiDAR, with growing work on fusing cameras to recover semantic cues that sparse point clouds lack~\cite{wangMultiModal3DObject2023}.
Detectors either operate on points, such as PointRCNN~\cite{shiPointRCNN3DObject2019}, or on a structured grid.
VoxelNet~\cite{zhouVoxelNetEndtoEndLearning2018} and its sparse-convolution successor SECOND~\cite{yanSECONDSparselyEmbedded2018} encode the scene as voxels.
PointPillars~\cite{langPointPillarsFastEncoders2019c} collapses the vertical axis into pillars to form a 2D pseudo-image, and PV-RCNN~\cite{shiPVRCNNPointVoxelFeature2020} combines a voxel backbone with point-based pooling.
CenterPoint~\cite{yinCenterbased3DObject2021} detects centers as keypoints in the bird's-eye view (BEV) and regresses box size and orientation without anchor hyperparameters.

Geometry alone is ambiguous where few points fall on an object.
BEVFusion~\cite{liuBEVFusionMultiTaskMultiSensor2023} lifts LiDAR and camera features into a shared BEV space.  TransFusion~\cite{baiTransFusionRobustLiDARCamera2022} replaces the geometric projection with a learned association between 3D queries and image features.
Both need large training sets.
Early fusion instead paints image evidence onto the point cloud, such that a standard LiDAR detector can be reused.
PointPainting~\cite{voraPointPaintingSequentialFusion2020} paints semantic class scores, PointAugmenting~\cite{wangPointAugmentingCrossModalAugmentation2021a} paints CNN features, and MVP~\cite{yinMultimodalVirtualPoint2021} adds virtual points on sparse far-range objects.

Trimodal LiDAR--RGB--thermal detection is far less common.
T360Fusion~\cite{tranT360FusionTemporal3602025} fuses LiDAR with 360$^\circ$ RGB and thermal panoramas in a shared BEV space, but it accumulates frames over time and, as a BEVFusion-based model, is computationally heavy.

\section{PROBLEM STATEMENT}
\label{sec:problem}

We consider 3D person detection from a rigidly mounted LiDAR, an RGB camera, and a thermal camera.
At frame~$t$, the observations are a LiDAR sweep $\mathbf{X} \in \mathbb{R}^{N \times 4}$, with returns $\mathbf{x}_n \in \mathbb{R}^4$ (coordinates and intensity), and images $I_c$ for $c \in \{\mathrm{rgb}, \mathrm{th}\}$.
The calibration defines the projection $(u_n, v_n) = \pi_c(\mathbf{x}_n)$ of each return into camera~$c$.

Each person is represented as an oriented box in the LiDAR frame,
\begin{equation}
  B_j = \bigl(\mathbf{c}_j,\, \mathbf{d}_j,\, \psi_j\bigr),
\label{eq:box-param}
\end{equation}
with center $\mathbf{c}_j \in \mathbb{R}^3$, size $\mathbf{d}_j = (l_j, w_j, h_j)$, and yaw $\psi_j \in (-\pi, \pi]$.
The ground truth is $\mathcal{B}_t^{\mathrm{gt}} = \{B_j^{\mathrm{gt}}\}_{j=1}^{K_t}$.
A detector $f_\theta$ predicts a scored set
\begin{equation}
\mathcal{B}_t^{\mathrm{pred}}
  = f_\theta\!\left(\mathbf{X},\, \{I_c\}_{c \in \{\mathrm{rgb}, \mathrm{th}\}}\right).
\label{eq:detector}
\end{equation}
Predictions are matched to $\mathcal{B}_t^{\mathrm{gt}}$ by 3D intersection-over-union, and performance is reported as average precision.

\section{METHOD}

This section describes the trimodal sensor-box, the calibration of the LiDAR, RGB, and thermal sensors, and the proposed early-fusion 3D detection network.

\begin{figure*}[!t]
  \centering
  \includegraphics[width=\textwidth]{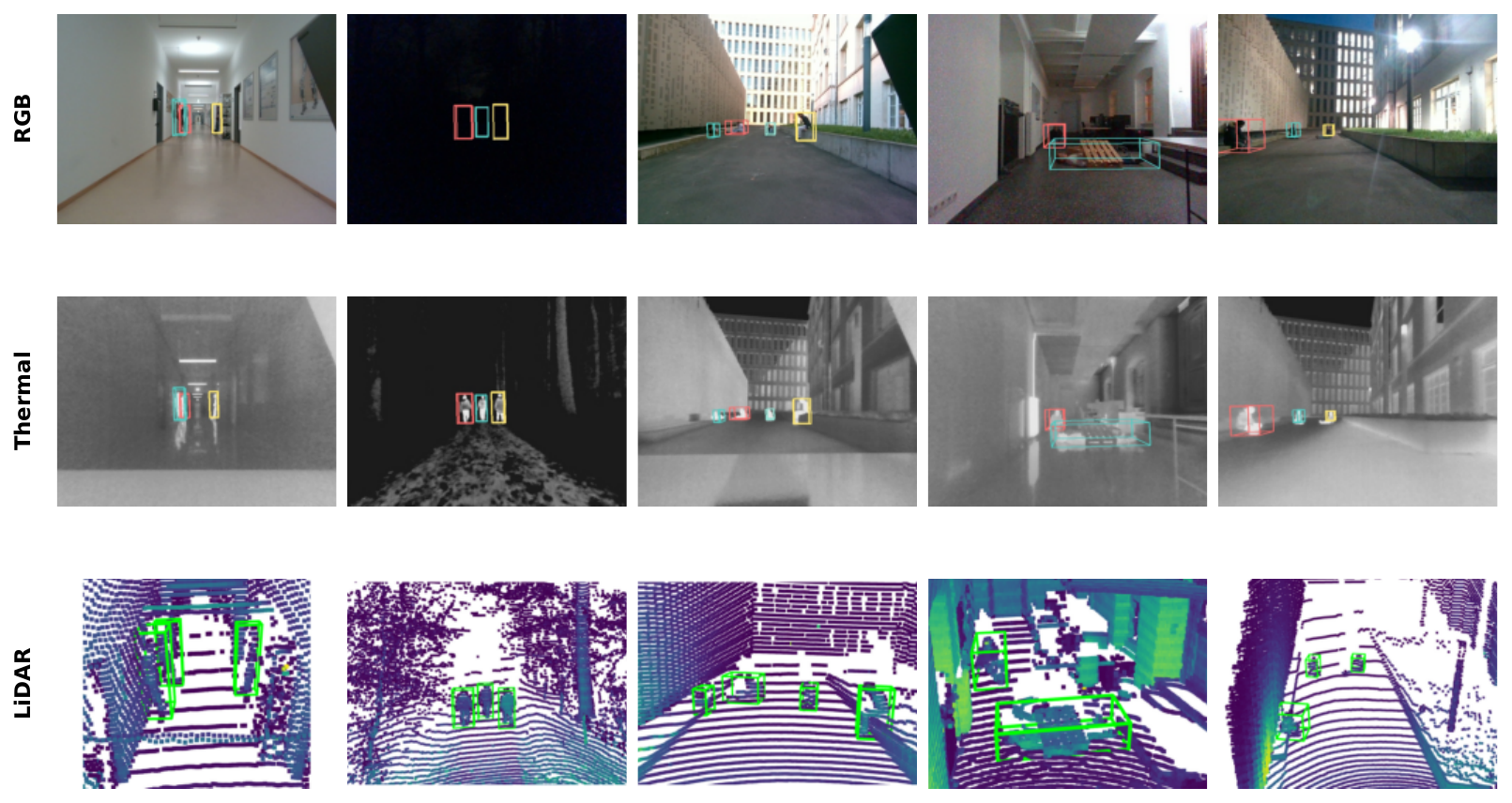}
  \caption{Representative dataset examples with 3D ground-truth bounding boxes on synchronized RGB images (top row), thermal images (middle row), and LiDAR point clouds (bottom row). From left to right: indoor hallway, outdoor forest at night, outdoor sitting pose, indoor lying person, and outdoor night scene.}
  \label{fig:dataset_samples}
\end{figure*}

\subsection{Sensor-box}
For data acquisition, we use a custom sensor-box that integrates an Ouster OS0 LiDAR, a FLIR A70 thermal camera, and an Intel RealSense D455 RGB-D camera (Fig.~\ref{fig:sensorbox}).

The LiDAR is mounted on top of the frame to minimize occlusions caused by the frame structure.
It provides a vertical field of view (FoV) of \SI{90}{\degree} with 128 channels (\SI{0.7}{\degree} angular resolution), a horizontal resolution of 1024 (\SI{0.35}{\degree}), a rotation rate of \SI{20}{Hz}, and a maximum range of \SI{250}{\meter}.

The RealSense D455 is mounted below the LiDAR. 
It captures RGB and depth images at $640 \times 480$ pixels with \SI{8}{bit} per channel. Only the RGB stream is used in this work.
The camera has a FoV of \SI{86}{\degree} $\times$ \SI{57}{\degree} and is configured to capture at \SI{15}{Hz}.

At the bottom, the FLIR A70 thermal camera is mounted. 
It captures at \SI{30}{Hz} with a resolution of $640 \times 480$ pixels and \SI{8}{bit} depth. It uses a \SI{95}{\degree} lens with a FoV of \SI{95}{\degree} $\times$ \SI{74}{\degree}.

All sensors are synchronized in software using timestamp-based alignment at \SI{10}{Hz}. This can induce minor motion artifacts in dynamic scenes. The error caused by the synchronization can be estimated using the velocity of the annotated tracked 3D instances and the time difference between the RGB/Thermal images and LiDAR. This results in a median error lower than \SI{10}{cm} for both RGB-LiDAR and Thermal-LiDAR.

\subsection{Calibration Target}
To calculate the extrinsic calibration between all three sensors, a new calibration target based on the target used in the Multi-Sensor Calibration Toolbox~\cite{rufMultiSensorCalibrationToolbox} is created.
The target is adapted to be visible in the thermal image in addition to the RGB image and the LiDAR point cloud.
To achieve this, ArUco markers are 3D printed from black and white PETG.
The white parts of each marker have a height of \SI{0.1}{mm} and the black parts have a height of \SI{3}{mm}.
A heating element is attached to the back of each marker, making them visible in the thermal and RGB images (Fig.~\ref{fig:calib_target}).
The holes in the target make it rotation-invariant and easily detectable in the LiDAR.
With this, the same target can be used for the extrinsic calibration between any two sensors.

For the intrinsic calibration of both the RGB and thermal camera, a checkerboard pattern is used.
The checkerboard pattern for the thermal camera is 3D printed from PETG with thin and thick squares and placed on a heated plate similar to the ArUco markers in the extrinsic calibration target.

\begin{figure*}[!t]
  \centering
  \includegraphics[width=\textwidth]{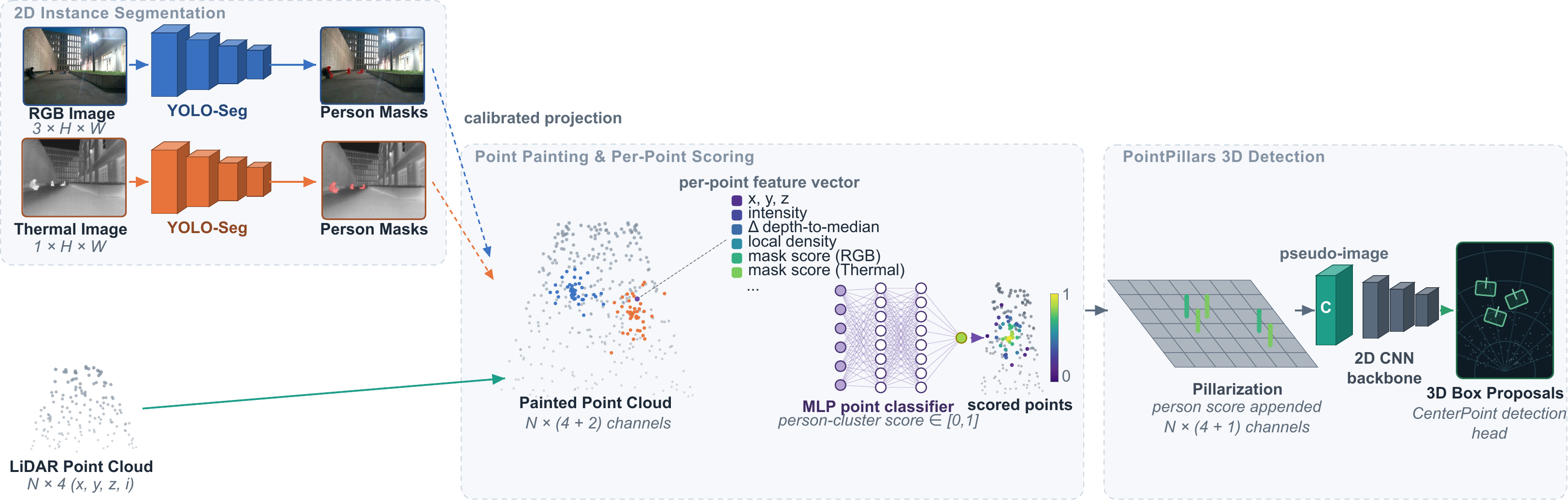}
  \caption{Overview of the proposed PaintFusion detection network. RGB and thermal images are processed by separate 2D instance segmentation streams. The resulting features are projected onto the LiDAR point cloud, scored per point, and fed into a PointPillars encoder with a CenterPoint detection head for 3D box prediction.}
  \label{fig:dnn_architecture}
\end{figure*}

\subsection{Calibration}
Using the new calibration target, the extrinsic calibration between the LiDAR and both cameras is initially calculated using the Multi-Sensor Calibration Toolbox~\cite{rufMultiSensorCalibrationToolbox}.
To further improve the calibration, an optimization-based algorithm is used.
This minimizes the re-projection error between each sensor and the calibration board for different poses of the calibration target using the Ceres solver~\cite{Agarwal_Ceres_Solver_2022}.
The images from the thermal and RGB cameras are undistorted before being used to calculate the calibration parameters.
The optimizer optimizes the RGB camera, thermal camera, and calibration board poses in the world frame $^w\mathbf{T}_{\mathrm{rgb}}, ^w\mathbf{T}_{\mathrm{th}}, \{{}^{w}\mathbf{T}_{b_i}\}_{i=1}^{N}$ for each board position $i$ as a joint nonlinear least-squares estimation problem.
To keep the system anchored, the LiDAR frame ($^w\mathbf{T}_l$) is set constant.

The corner $^b\mathbf{P}_{ij}$, with index $j$ as the marker corner, is transformed into camera frame $c\!\in\!\{\mathrm{rgb},\mathrm{th}\}$ using the poses $^w\mathbf{T}_c$ and $^w\mathbf{T}_{b_i}$:
\begin{equation}
    ^c\mathbf{P}_{ij} = {^w\mathbf{T}_c}^{-1} \cdot ^w\mathbf{T}_{b_i} \cdot ^b\mathbf{P}_{ij}.
\end{equation}
The pinhole intrinsics $f_x,f_y,c_x,c_y$ project this corner $(X,Y,Z,1)^\top$ to the predicted pixel $\mathbf{\hat{z}}_{ij}$:
\begin{equation}
    \mathbf{\hat{z}}_{ij} =
\begin{bmatrix}
f_x \dfrac{X}{Z} + c_x \\
f_y \dfrac{Y}{Z} + c_y
\end{bmatrix}.
\end{equation}
The camera residual compares that prediction with the observed pixel $\mathbf{z}_{ij}$, normalized by $\sigma_c=\SI{1.0}{px}$:
\begin{equation}
\mathbf{r}^{\mathrm{c}}_{ij} = \frac{\mathbf{\hat{z}}_{ij} -\mathbf{z}_{ij}}{\sigma_c}.
\end{equation}
The LiDAR residual is calculated by expressing the corners of the board and the corners detected with the LiDAR $^l\mathbf{P}_{ij}$ using ICP into world frame
\begin{align}
    ^w\mathbf{P}_{ij}^{(b)} &= ^w\mathbf{T}_{b_i} \cdot ^b\mathbf{P}_{ij} \\
    ^w\mathbf{P}_{ij}^{(l)} &= ^w\mathbf{T}_l \cdot ^l\mathbf{P}_{ij}.
\end{align}
Their difference, normalized by $\sigma_l=\SI{0.02}{\meter}$ is used as the residual:
\begin{equation}
\mathbf{r}^{\mathrm{l}}_{ij} = \frac{^w\mathbf{P}_{ij}^{(l)} - ^w\mathbf{P}_{ij}^{(b)}}{\sigma_l}.
\end{equation}
The solution is then obtained by minimizing the weighted sum of squared residuals, with $\rho(\cdot)$ the Huber loss and the parameters $\Theta=\{^w\mathbf{T}_{\mathrm{rgb}},^w\mathbf{T}_{\mathrm{th}},\{{}^{w}\mathbf{T}_{b_i}\}_{i=1}^{N}\}$:
\begin{equation}
\begin{aligned}
\Theta^{*}
&=
\arg\min_{\Theta}
\Bigg(
\sum_{i,j}
\rho\!\left(\left\|
\mathbf{r}^{\mathrm{rgb}}_{ij}
\right\|^2\right)
+
\sum_{i,j}
\rho\!\left(\left\|
\mathbf{r}^{\mathrm{th}}_{ij}
\right\|^2\right)
\\
&\qquad\qquad
+
\sum_{i,j}
\rho\!\left(\left\|
\mathbf{r}^{\mathrm{l}}_{ij}
\right\|^2\right)
\Bigg).
\end{aligned}
\end{equation}

To evaluate the performance of the calibration, the average camera-camera and camera-LiDAR calibration errors are calculated over five different recordings with the calibration target covering large parts of the image.
This results in a camera-camera error of \SI{1.56}{px}, an RGB-LiDAR error of \SI{3.58}{px}, and a thermal-LiDAR error of \SI{2.96}{px}.
\begin{figure*}[!t]
    \centering
    \definecolor{bootstrapours}{HTML}{4E79A7}
    \newsavebox{\mapcomparetablebox}
    \newsavebox{\mapcompareplotbox}

    \savebox{\mapcomparetablebox}{%
        \setlength{\tabcolsep}{3pt}%
        \renewcommand{\arraystretch}{1.05}%
        \footnotesize
        \begin{tabular}{@{}llcccccc@{}}
            \toprule
            \textbf{Regime} & \textbf{Method}
                & \textbf{AP@50}
                & \textbf{Standing}
                & \textbf{Lying}
                & \textbf{Walking}
                & \textbf{Sitting}
                & \textbf{\shortstack{Latency\\{[}\si{ms}{]}}} \\
            \midrule
            \multirow{4}{*}{LiDAR-only}
                & PointPillars~\cite{langPointPillarsFastEncoders2019c} & 47.1 & 20.6 & 11.0 & 59.8 & 27.1 & \textbf{15.6} \\
                & PV-RCNN~\cite{shiPVRCNNPointVoxelFeature2020}         & 38.6 & 20.2 & 12.0 & 44.1 & 25.3 & 79.6 \\
                & SECOND~\cite{yanSECONDSparselyEmbedded2018}           & 32.6 & 18.0 & 22.1 & 41.2 & 14.2 & 23.8 \\
                & CenterPoint~\cite{yinCenterbased3DObject2021}         & 53.4 & 21.9 & \textbf{23.8} & 62.5 & 33.9 & 31.5 \\
            \midrule
            \multirow{2}{*}{RGB+LiDAR}
                & BEVFusion~\cite{liuBEVFusionMultiTaskMultiSensor2023} & 60.5 & 32.3 & 22.3 & 62.6 & 45.0 & 101.0 \\
                & PointPainting~\cite{voraPointPaintingSequentialFusion2020} & 43.7 & 23.9 & 1.6 & 55.1 & 15.1 & 57.7 \\
            \midrule
            Thermal+LiDAR
                & Thermal PointPainting & 52.2 & 32.9 & 2.9 & 62.3 & 22.8 & 56.9 \\
            \midrule
            \multirow{3}{*}{Trimodal}
                & Tri-BEVFusion                                           & 60.7 & \textbf{36.4} & 19.3 & \textbf{71.0} & 38.8 & 107.1 \\
                & Tri-PointPainting                                       & 51.4 & 21.7 & 3.9 & 64.7 & 19.1 & 105.6 \\
                & \textbf{PaintFusion (Ours)}                                                    & \textbf{65.4} & 22.6 & 18.4 & 68.0 & \textbf{49.7} & 51.6 \\
            \bottomrule
        \end{tabular}%
    }

    \savebox{\mapcompareplotbox}{%
        \begin{tikzpicture}
            \begin{axis}[
                scale only axis,
                width=0.30\textwidth,
                height=5.5cm,
                xmin=20,
                xmax=80,
                xtick={20,40,60,80},
                ymin=0.4,
                ymax=10.7,
                xlabel={AP@50 [\%]},
                ytick=\empty,
                y dir=reverse,
                xmajorgrids=true,
                grid style={gray!22},
                tick label style={font=\footnotesize},
                label style={font=\footnotesize},
                clip=false,
                enlarge y limits=0.02,
            ]
                \addplot[
                    only marks,
                    mark=*,
                    mark size=1.5pt,
                    color=black!55,
                    error bars/.cd,
                    x dir=both,
                    x explicit,
                ] coordinates {
                    (59.6, 2) +- (14.7, 14.0)
                    (59.1, 3) +- (14.6, 12.8)
                    (51.2, 4) +- (9.1, 14.7)
                    (47.6, 5) +- (12.3, 13.3)
                    (46.9, 6) +- (11.5, 13.4)
                    (41.2, 7) +- (9.8, 10.6)
                    (37.5, 8) +- (11.0, 13.2)
                    (32.7, 9) +- (8.5, 9.4)
                    (29.7, 10) +- (9.1, 11.9)
                };
                \addplot[
                    only marks,
                    mark=*,
                    mark size=1.8pt,
                    color=bootstrapours,
                    error bars/.cd,
                    x dir=both,
                    x explicit,
                ] coordinates {
                    (62.6, 1) +- (13.8, 10.9)
                };
                \node[anchor=south west, font=\scriptsize\rmfamily\bfseries, text=bootstrapours] at (axis cs:48.8,1) {Ours};
                \node[anchor=south west, font=\scriptsize] at (axis cs:44.9,2) {BEVFusion};
                \node[anchor=south west, font=\scriptsize] at (axis cs:44.5,3) {Tri-BEVFusion};
                \node[anchor=south west, font=\scriptsize] at (axis cs:42.1,4) {CenterPoint};
                \node[anchor=south west, font=\scriptsize] at (axis cs:35.3,5) {Thermal PointPainting};
                \node[anchor=south west, font=\scriptsize] at (axis cs:35.4,6) {Tri-PointPainting};
                \node[anchor=south west, font=\scriptsize] at (axis cs:31.4,7) {PointPillars};
                \node[anchor=south west, font=\scriptsize] at (axis cs:26.5,8) {PointPainting};
                \node[anchor=south west, font=\scriptsize] at (axis cs:24.2,9) {PV-RCNN};
                \node[anchor=south west, font=\scriptsize] at (axis cs:20.6,10) {SECOND};
            \end{axis}
        \end{tikzpicture}%
    }

    \noindent
    \makebox[0.60\textwidth][l]{\raisebox{\dp\mapcomparetablebox}{\usebox{\mapcomparetablebox}}}%
    \hfill
    \makebox[0.36\textwidth][c]{\raisebox{\dimexpr\dp\mapcompareplotbox-0.7cm\relax}{\usebox{\mapcompareplotbox}}}

    \vspace{0.4em}
    \noindent
    \begin{minipage}[t]{0.50\textwidth}
        \captionof{table}{Comparison of AP@50 across detection methods, sensor regimes, and pose tags.}
        \label{tab:map_comparison}
    \end{minipage}\hfill
    \begin{minipage}[t]{0.36\textwidth}
        \captionof{figure}{95\% bootstrap confidence intervals for overall AP@50 on the test split}
        \label{fig:bootstrap_map_ci}
    \end{minipage}
\end{figure*}

\subsection{Dataset Annotation}
Each person is annotated with an oriented 3D bounding box, a pose tag, and an instance mask in the RGB and thermal image.
The 3D boxes are obtained from the point cloud.
After background removal, HDBSCAN~\cite{10.1007/978-3-642-37456-2_14} forms clusters.
Only clusters whose projection overlaps a detection in either image are kept.
A Kalman filter tracks the accepted clusters.
The resulting 3D boxes and pose tags are corrected in CVAT~\cite{boris_sekachev_2020_4009388}.
Each corrected 3D box is projected into both images and assigned to one detection by Hungarian matching.
The matched instance mask is kept as the 2D annotation.
Detections without a corrected 3D box are discarded.
\subsection{Detection Network}
\label{sec:detection}
For localizing people in 3D we use an early-fusion pipeline as depicted in Fig.~\ref{fig:dnn_architecture}.
It divides the detection process into three
stages: (i)~2D instance segmentation on the RGB and
thermal images, (ii)~painting of mask evidence onto the LiDAR point cloud and
per-point scoring, and (iii)~3D box proposal generation from the decorated point cloud.

For 2D instance segmentation we use the YOLO26~\cite{jocherUltralyticsYOLO26Unified2026} model in its
lightweight yolo26n-seg version, fine-tuned separately on the RGB and thermal masks.
This enables parallel processing of both
camera streams.
Each detector instance consists of a binary person mask~$M_{c,k}$ with confidence
$s_{c,k}$ on image~$I_c$.
LiDAR points where the projection $(u_n,v_n)=\pi_c(\mathbf{x}_n)$ falls inside any mask become candidate points:
\begin{equation}
  \mathcal{S}_{c,k}
  = \bigl\{\, n \;\big|\; M_{c,k}(u_n, v_n) = 1 \,\bigr\}.
  \label{eq:seed-pool}
\end{equation}
Because both cameras project the same sweep~$\mathbf{X}$, RGB and thermal
evidence are fused at LiDAR index~$n$ to distinguish between foreground and background points. 
This ensures that even if one camera's detection fails, the pipeline can still
use the detection information from the other camera.

Each candidate point is encoded by a 33-dimensional feature vector. These features capture
depth and geometry relative to its instance (depth rank, nearest depth
cluster, height above ground), mask layout (normalized position, boundary
distance, area), local seed support, LiDAR intensity, cross-instance
overlap, detector confidence~$s_{c,k}$ as well as cross-camera agreement on the
same return.

A feed-forward MLP ($512\!\to\!256\!\to\!128$, ReLU, dropout $p{=}0.1$) maps the 
preprocessed features~$\tilde{\mathbf{f}}_n$ to a person probability
$\hat{p}_n = \sigma(g(\tilde{\mathbf{f}}_n))$.
Training minimizes a binary cross-entropy against labels $y_n \in \{0,1\}$, where a candidate is positive when its LiDAR return lies inside a ground-truth person box of the evaluation set in Sec.~\ref{sec:eval_setup}.

In the final stage, each LiDAR point is decorated with its MLP score
$\hat{p}_n$. This adds a new channel to the point cloud.
PointPillars~\cite{langPointPillarsFastEncoders2019c} groups the decorated
points into vertical pillars and projects extracted features into a bird's-eye-view
pseudo-image. After processing the image with a convolutional backbone, a CenterPoint \cite{yinCenterbased3DObject2021} detection head regresses oriented 3D box proposals
$B_j = (\mathbf{c}_j, \mathbf{d}_j, \psi_j)$ with confidence scores, producing the final predicted set~$\mathcal{B}_t^{\mathrm{pred}}$.


\section{EVALUATION}

This section presents the evaluation of our trimodal detection pipeline on TriSAR-3D against LiDAR-only and multimodal baselines.

\subsection{Experimental Setup}
\label{sec:eval_setup}
We split the annotated frames into training (80\%), validation (10\%), and test (10\%) sets. As individual sequences are too long to be assigned to a single split without leaving sets unrepresentative of all locations, poses, and lighting conditions, we partition each sequence into subsequences of \SIrange{10}{50}{s} and assign these to the sets. To mitigate temporal correlation, we enforce a buffer of \SI{1}{s} between subsequences assigned to different splits and exclude the frames within it. 
During training we use random rotation, translation, and scale augmentations on the data.
The evaluation is restricted to people with their 3D box center within \SIrange{5}{30}{m} and excludes severely occluded instances with less than 5 points inside the 3D box or if at least 50\% of its 3D box, projected into the RGB image, is covered by a closer person. Instances that are only partially occluded, for example instances where a person is buried by debris (as in Fig.~\ref{fig:dataset_samples} column 4), are still included. This makes sure that people are visible in both camera frames and have a minimum number of LiDAR points in their 3D box. 
In addition to the overall AP@50 score, we report pose-conditioned AP@50 scores by filtering both predictions and ground truth to instances tagged as standing, lying, walking, or sitting. The latency is measured per frame on a laptop with an Intel Core i9-12900H and NVIDIA GeForce RTX~3080 Ti Laptop GPU and is averaged over the test set.

We compare four different sensor setups: LiDAR-only, RGB+LiDAR, Thermal+LiDAR and trimodal (RGB+thermal+LiDAR). As LiDAR-only baselines we use PointPillars~\cite{langPointPillarsFastEncoders2019c}, SECOND~\cite{yanSECONDSparselyEmbedded2018}, PV-RCNN~\cite{shiPVRCNNPointVoxelFeature2020}, and CenterPoint~\cite{yinCenterbased3DObject2021}.
For RGB+LiDAR fusion we test the BEVFusion~\cite{liuBEVFusionMultiTaskMultiSensor2023} and PointPainting~\cite{voraPointPaintingSequentialFusion2020} architectures. All baselines except PointPainting use the implementations provided by mmdetection3d~\cite{mmdet3d2020}. PointPainting uses a DeepLabv3+ network to generate the segmentation scores followed by SECOND as the LiDAR decoder.
To isolate the effect of thermal sensing, we additionally implement thermal PointPainting and trimodal variants of these pipelines (Tri-BEVFusion and Tri-PointPainting) by adding a second image stream. For Tri-PointPainting we append another channel with the class score from the thermal image to the processed point cloud. 
For Tri-BEVFusion both the thermal and RGB images are used as the Multi-View RGB images input to the models with the same pretrained image backbone without additional pretraining. All models are trained on the new dataset using the same distance and occlusion limitations. 

\begin{table*}[!t]
    \centering
    \caption{Impact of removing \emph{lying} instances from training on AP@50. Each entry lists performance for full training / training without \emph{lying} instances}
    \label{tab:lying_ablation}
    \begin{tabular}{llccccc}
        \toprule
        \textbf{Regime} & \textbf{Method}
            & \textbf{Overall}
            & \multicolumn{4}{c}{\textbf{AP@50 by pose}} \\
        \cmidrule(lr){4-7}
        & & & \textbf{Standing} & \textbf{Lying} & \textbf{Walking} & \textbf{Sitting} \\
        \midrule
        \multirow{4}{*}{LiDAR-only}
            & PointPillars~\cite{langPointPillarsFastEncoders2019c} & 47.1 / 40.6 & 20.6 / 8.3 & 11.0 / 0.1 & 59.8 / 59.9 & 27.1 / 19.6 \\
            & PV-RCNN~\cite{shiPVRCNNPointVoxelFeature2020}         & 38.6 / 39.7 & 20.2 / 13.8 & 12.0 / \textbf{0.3} & 44.1 / 47.7 & 25.3 / 31.6 \\
            & SECOND~\cite{yanSECONDSparselyEmbedded2018}           & 32.6 / 28.3 & 18.0 / 5.2 & 22.1 / 0.0 & 41.2 / 39.5 & 14.2 / 19.1 \\
            & CenterPoint~\cite{yinCenterbased3DObject2021}          & 53.4 / 49.4 & 21.9 / 11.1 & \textbf{23.8} / 0.1 & 62.5 / 62.6 & 33.9 / 37.0 \\
        \midrule
        \multirow{2}{*}{RGB+LiDAR}
            & BEVFusion~\cite{liuBEVFusionMultiTaskMultiSensor2023} & 60.5 / 54.8 & 32.3 / 25.6 & 22.3 / 0.2 & 62.6 / \textbf{71.0} & 45.0 / 32.9 \\
            & PointPainting~\cite{voraPointPaintingSequentialFusion2020} & 43.7 / 41.1 & 23.9 / \textbf{40.1} & 1.6 / 0.0 & 55.1 / 61.3 & 15.1 / 10.1 \\
        \midrule
        Thermal+LiDAR
            & Thermal PointPainting & 52.2 / 39.7 & 32.9 / 17.3 & 2.9 / 0.0 & 62.3 / 56.5 & 22.8 / 5.2 \\
        \midrule
        \multirow{3}{*}{Trimodal}
            & Tri-BEVFusion                                           & 60.7 / 53.7 & \textbf{36.4} / 36.3 & 19.3 / 0.2 & \textbf{71.0} / 69.4 & 38.8 / 32.2 \\
            & Tri-PointPainting                                       & 51.4 / 45.0 & 21.7 / 24.8 & 3.9 / 0.0 & 64.7 / 63.1 & 19.1 / 14.3 \\
            & \textbf{PaintFusion (Ours)}                                                    & \textbf{65.4} / \textbf{62.4} & 22.6 / 32.6 & 18.4 / 0.2 & 68.0 / 69.3 & \textbf{49.7} / \textbf{52.0} \\
        \bottomrule
    \end{tabular}
\end{table*}

\subsection{Overall Detection Performance}
We compare 3D person detection performance across the four sensor regimes in \Tabref{tab:map_comparison}.
Fig.~\ref{fig:bootstrap_map_ci} reports 95\% confidence intervals from a cluster bootstrap over test subsequences ($N{=}1000$ resamples).
Among LiDAR-only detectors, CenterPoint achieves the strongest overall AP@50 of 53.4, followed by PointPillars at 47.1.
PV-RCNN with 38.6 and SECOND with 32.6 lag behind. CenterPoint's anchor-free center regression imposes no size or orientation prior. This makes it easier for the network to adapt to the varying geometries of sitting and lying persons. All LiDAR-only methods are faster than the fusion-based architectures with the exception of PV-RCNN, owing to its two-stage design.

Adding RGB information yields a clear improvement.
BEVFusion reaches an overall AP@50 of 60.5, which is an improvement of 7.1 AP over the best LiDAR-only method.
These results confirm that semantic cues from a color camera can help disambiguate persons when geometric evidence alone is weak.
PointPainting performs better than SECOND and PV-RCNN. However, compared to CenterPoint, it performs worse. This is consistent with reporting from the literature \cite{wangPointAugmentingCrossModalAugmentation2021a}. Due to the need to process camera information as well, the latency of the detectors increases. 

Using thermal information instead of additional RGB information yields a further improvement from PointPainting by 8.5 AP over RGB, suggesting that thermal information is more useful than RGB information independent of pose.

The trimodal regime further improves performance.
Our method achieves the highest overall AP@50 of 65.4 and the best sitting score of 49.7.
Adding a thermal image stream to PointPainting yields an improvement of 7.7 AP over the RGB-only PointPainting baseline. This shows that such information can provide a benefit even without architectural changes. Tri-BEVFusion reaches the highest walking AP@50 of 71.0. While Tri-PointPainting and Tri-BEVFusion show the highest latency over all regimes, with \SI{105.6}{ms} and \SI{107.1}{ms} respectively, our method runs at a substantially lower latency of \SI{51.6}{ms} per frame. 

\subsection{Impact of Pose Diversity in Training}

Rescue scenarios require detectors that generalize to poses rarely seen in automotive datasets.
To quantify how sensitive learning-based detectors are to training-set composition, we ablate the \emph{lying} category by training each method either on the full dataset or on a subset from which all lying instances are removed.
\Tabref{tab:lying_ablation} reports AP@50 under both training conditions on the full test split and on each pose tag.
The overall AP@50 drops only modestly for most methods, but the lying-specific AP collapses almost entirely. This drop is independent of the architecture used. It suggests that training-set diversity of the TriSAR-3D dataset exerts a far larger effect on the performance of the detector than the specific architecture used. This highlights the need for datasets with diverse poses.

\begingroup
\setlength{\textfloatsep}{6pt plus 2pt minus 2pt}
\begin{figure}[!t]
    \centering
    \begin{tikzpicture}
        \begin{axis}[
            name=mapgain,
            scale only axis,
            width=0.72\columnwidth,
            height=0.62\columnwidth,
            ybar,
            bar width=0.50cm,
            xmin=0.5,
            xmax=5.5,
            xtick={1,2,3,4,5},
            xticklabels={5--10, 10--15, 15--20, 20--25, 25--30},
            xlabel={Distance bin [\si{m}]},
            ylabel={Absolute AP@50 gain [AP]},
            ymin=0,
            ymax=25,
            ytick={0,5,10,15,20,25},
            nodes near coords={\pgfmathprintnumber[fixed, precision=1]{\pgfplotspointmeta}},
            nodes near coords align={vertical},
            every node near coord/.append style={
                font=\scriptsize,
                anchor=south,
                yshift=1pt,
            },
            legend style={
                at={(0.5,1.04)},
                anchor=south,
                legend columns=1,
                draw=none,
                fill=none,
                font=\footnotesize
            },
            single ybar legend,
        ]
            \addplot[
                fill=black!18,
                draw=black,
                bar shift=-0.27cm,
            ] coordinates {
                (1, 17.3)
                (2, 22.3)
                (3, 23.2)
                (4, 16.6)
                (5, 19.6)
            };
            \addlegendentry{Ours vs.\ mean LiDAR-only}
            \addplot[
                fill=black!45,
                draw=black,
                bar shift=0.27cm,
            ] coordinates {
                (1, 0.7)
                (2, 19.6)
                (3, 11.4)
                (4, 8.1)
                (5, 15.8)
            };
            \addlegendentry{Mean RGB+LiDAR vs.\ mean LiDAR-only}
            \addlegendimage{
                legend image code/.code={
                    \draw[yshift=-0.2em, fill=black, draw=black] (3pt,0.4em) circle [radius=0.9pt];
                },
            }
            \addlegendentry{Avg.\ points / instance}
        \end{axis}
        \coordinate (mapgainplotsw) at (current bounding box.south west);
        \begin{axis}[
            scale only axis,
            width=0.72\columnwidth,
            height=0.62\columnwidth,
            at={(mapgain.south west)},
            anchor=south west,
            hide x axis,
            axis y line*=right,
            xmin=0.5,
            xmax=5.5,
            ymode=log,
            log basis y={10},
            ymin=8,
            ymax=1200,
            ytick={10,100,1000},
            ylabel={Avg.\ LiDAR points / instance (log)},
        ]
            \addplot[
                black,
                thick,
                mark=*,
                mark size=1.8pt,
                nodes near coords,
                point meta=explicit symbolic,
                every node near coord/.append style={
                    font=\scriptsize,
                    anchor=south,
                    yshift=6pt,
                    xshift=-0.27cm,
                },
            ] coordinates {
                (1, 134.8) [135]
                (2, 40.3) [40]
                (3, 23.1) [23]
                (4, 15.0) [15]
                (5, 8.5) [9]
            };
        \end{axis}
        \coordinate (mapgainplotne) at (current bounding box.north east);
        \pgfresetboundingbox
        \useasboundingbox (mapgainplotsw) rectangle (mapgainplotne);
    \end{tikzpicture}
    \vspace{-0.15cm}
    \caption{Absolute AP@50 gain of our method and the two LiDAR-RGB fusion baselines over the mean of four LiDAR-only methods  (bars, left axis) and mean LiDAR points per person instance (right axis).}
    \label{fig:map_gain_distance}
\end{figure}
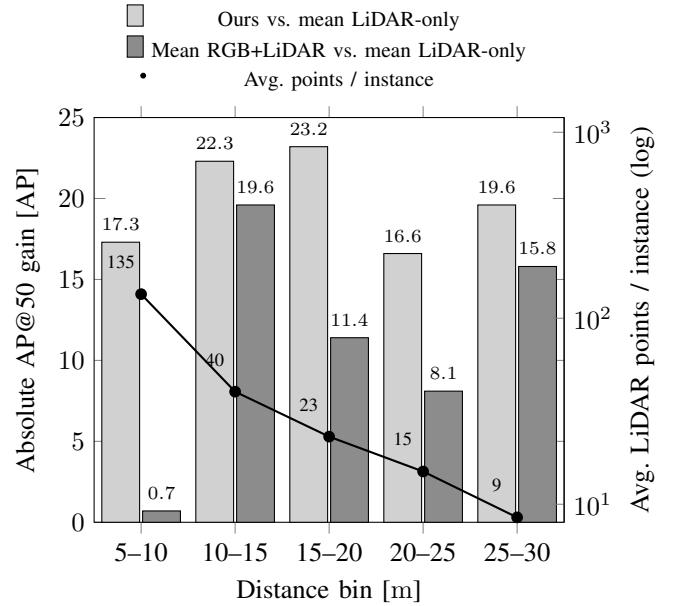
\endgroup

\subsection{Distance-Stratified Analysis}
Search-and-rescue robots must detect victims at varying ranges, from close indoor encounters to persons tens of meters away in open terrain.
Fig.~\ref{fig:map_gain_distance} reports the absolute AP@50 gain over the mean of four LiDAR-only baselines (PointPillars, PV-RCNN, SECOND, and CenterPoint) for our trimodal pipeline and the gain of the mean of the two RGB+LiDAR fusion methods against the mean of the LiDAR-only baselines. We observe that the number of LiDAR points per instance decreases significantly over distance from an average of 135 points in the range \SIrange{5}{10}{m} to only 9 points in the range \SIrange{25}{30}{m} despite the high-resolution LiDAR. Our trimodal method consistently improves over the LiDAR-only mean by \SIrange{17.3}{23.2}{AP} across all distance bins. The RGB+LiDAR fusion mean shows a smaller and more variable gain (+0.7 to +19.6~AP). This strengthens the argument that camera-based information helps especially where geometric evidence alone is weak. When only a few LiDAR points are projected on a person, it is difficult to tell that person apart from other objects and camera-based semantic information helps to disambiguate these cases. Our method further improves over the RGB+LiDAR mean by up to 11.8~AP, indicating that thermal sensing provides complementary cues beyond RGB alone.

\subsection{Limitations}
Our study has several limitations.
All recordings use a fixed sensor rig rather than a moving robot platform. Thus the images do not contain ego-motion-induced motion blur, similar to the TriStar~\cite{stippelTrimodalDatasetRGB2024a} and R-LiViT~\cite{mirlachRLiViTLiDARVisualThermalDataset2025} datasets. Moreover, environmental conditions such as fog, dust, or smoke commonly present in search-and-rescue settings are currently not included in the dataset. Both are left as important future work.

\section{CONCLUSION AND OUTLOOK}

We introduced \emph{TriSAR-3D}, a trimodal dataset with synchronized 128-channel LiDAR, RGB, and thermal streams, recorded indoors and outdoors and annotated with 3D bounding boxes, 2D segmentation masks, and pose tags.
Compared with existing trimodal datasets, it offers a higher LiDAR resolution and covers different environments with explicit pose labels.

PaintFusion combines lightweight 2D segmentation, MLP-based point scoring, and a PointPillars encoder with a CenterPoint head.
It reaches the best overall AP@50 of the methods we evaluate, at roughly half the latency of the other multimodal baselines.
Adding thermal sensing helps when RGB is already present, and multimodal fusion helps most where few LiDAR points fall on a person, for example at long range.
Removing lying instances from training collapses the lying-pose AP across all architectures, highlighting that the composition of the training set matters more than the choice of detector.

We plan to extend the dataset with dust, fog, and smoke, and to mount the sensor-box on a mobile robot so that ego-motion and onboard runtime are included.
On the method side, we plan to add tracking across frames to stabilize detections in clutter and improve recall on partially observable persons. 

\bibliographystyle{IEEEtran}
\bibliography{bibliography}

\end{document}